\documentclass[sigconf]{acmart}
\usepackage{multirow}
\AtBeginDocument{%
  }

\copyrightyear{2025}
\acmYear{2025}
\setcopyright{acmlicensed}\acmConference[RichMediaGAI '25]{Proceedings of the 3rd International Workshop on Rich Media With Generative AI}{October 27--28, 2025}{Dublin, Ireland}
\acmBooktitle{Proceedings of the 3rd International Workshop on Rich Media With Generative AI (RichMediaGAI '25), October 27--28, 2025, Dublin, Ireland}
\acmDOI{10.1145/3746262.3761973}
\acmISBN{979-8-4007-2044-4/2025/10}
\begin{document}

\title{SLMQuant: Benchmarking Small Language Model Quantization for Practical Deployment}

\author{Jiacheng Wang\textsuperscript{*}}
\affiliation{%
 \institution{SKLCCSE}
 \institution{School of Artificial Intelligence}
 \institution{Beihang University}
  \city{Beijing}
  \country{China}}
  \email{21371478@buaa.edu.cn}

\author{Yejun Zeng\textsuperscript{*}}
\affiliation{%
 \institution{SKLCCSE}
 \institution{School of Artificial Intelligence}
 \institution{Beihang University}
  \city{Beijing}
  \country{China}}
  \email{zengyejun@buaa.edu.cn}

\author{Jinyang Guo
\textsuperscript{†}
}
\affiliation{%
 \institution{SKLCCSE}
 \institution{School of Artificial Intelligence}
 \institution{Beihang University}
  \city{Beijing}
  \country{China}}
\email{jinyangguo@buaa.edu.cn}
 
\author{Yuqing Ma}
\affiliation{%
 \institution{SKLCCSE}
 \institution{School of Artificial Intelligence}
 \institution{Beihang University}
  \city{Beijing}
  \country{China}}
\email{yuqingma@buaa.edu.cn}

\author{Aishan Liu}
\affiliation{%
 \institution{SKLCCSE}
 \institution{School of Computer Science and Engineering}
 \institution{Beihang University}
  \city{Beijing}
  \country{China}}
\email{liuaishan@buaa.edu.cn}

\author{Xianglong Liu}
\affiliation{%
 \institution{SKLCCSE}
 \institution{School of Computer Science and Engineering}
 \institution{Beihang University}
  \city{Beijing}
  \country{China}}
\email{xlliu@buaa.edu.cn}

\renewcommand{\shortauthors}{Jiacheng Wang et al.}

\begin{abstract}
  Despite the growing interest in Small Language Models (SLMs) as resource-efficient alternatives to Large Language Models (LLMs), their deployment on edge devices remains challenging due to unresolved efficiency gaps in model compression. While quantization has proven effective for LLMs, its applicability to SLMs is significantly underexplored, with critical questions about differing quantization bottlenecks and efficiency profiles. This paper introduces SLMQuant, the first systematic benchmark for evaluating LLM compression techniques when applied to SLMs. Through comprehensive multi-track evaluations across diverse architectures and tasks, we analyze how state-of-the-art quantization methods perform on SLMs. Our findings reveal fundamental disparities between SLMs and LLMs in quantization sensitivity, demonstrating that direct transfer of LLM-optimized techniques leads to suboptimal results due to SLMs' unique architectural characteristics and training dynamics. We identify key factors governing effective SLM quantization and propose actionable design principles for SLM-tailored compression. SLMQuant establishes a foundational framework for advancing efficient SLM deployment on low-end devices in edge applications, and provides critical insights for deploying lightweight language models in resource-constrained scenarios.
\end{abstract}




\pdfstringdefDisableCommands{\def\z@{}}
\maketitle
 \renewcommand{\thefootnote}{\fnsymbol{footnote}}
 \setcounter{footnote}{0}
 \footnotetext[1]{These authors contributed equally.}
 \footnotetext[2]{Corresponding author: jinyangguo@buaa.edu.cn}

\section{Introduction}
In recent years, large language models (LLMs) have garnered considerable interest due to their superior performance across a wide range of natural language tasks. However, their deployment in real-world applications remains challenging due to the substantial computational and storage demands inherent to their scale. To mitigate these limitations, small language models (SLMs)~\cite{allal2025smollm2, hui2024qwen2, team2023gemini,liu2024lta-pcs,he2025da-kd}, have been proposed as lightweight alternatives. While SLMs demonstrate notable reductions in parameter size and computational overhead relative to their larger counterparts, they still exhibit significant efficiency gaps that hinder their deployment on edge devices with limited computational resources. 

\begin{figure}[!t]
  \centering
  \includegraphics[width=0.9\linewidth]{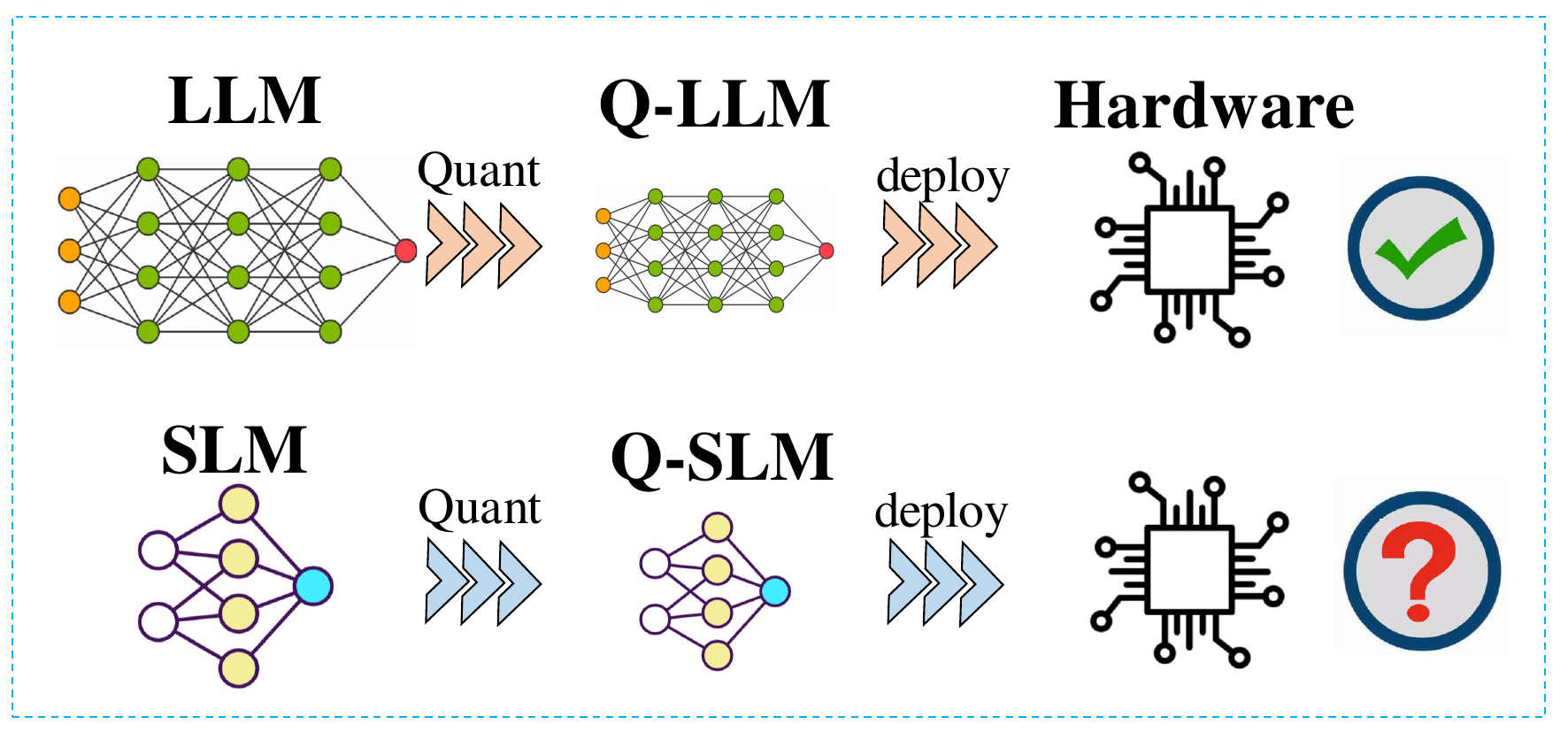} 
    \caption{The motivation of SLMQuant. }
  \label{fig:1}
\end{figure}

To mitigate this gap, model compression techniques~\cite{guo2020multi, guo2023multidimensional, guo2020channel, guo2020model}have been extensively investigated to enable practical deployment of LLMs on resource-constrained devices. Among these, quantization has emerged as a prominent strategy, aiming to replace full-precision model parameters with low-precision representations to reduce computational overhead and memory footprint. While prior quantization methods have predominantly focused on LLMs~\cite{gong2024llmc}\cite{guo2024compressing}\cite{lv2024ptq4sam}, their applicability to SLMs remains underexplored. Recent studies indicate that SLMs, despite their compact architectures, often exhibit unique sensitivity to quantization-induced precision loss due to their limited parameter capacity and distinct training dynamics. This raises critical challenges in developing quantization frameworks that preserve performance while achieving aggressive compression. To date, systematic investigations into quantization strategies for SLMs have received insufficient attention. Addressing this open problem is essential for advancing the deployment of SLMs in resource-limited scenarios.


Designing effective quantization algorithms for SLMs presents two central challenges. The first lies in identifying the critical bottleneck for quantization in SLM architectures. Prior work on LLMs has established that quantization bottlenecks stem from the outliers, which disproportionately degrade performance under low-precision constraints~\cite{xiao2023smoothquant,guo2021jointpruning,guo20223d,guo2023cbanet}. These outliers are attributed to the training paradigms of LLMs~\cite{touvron2023llama}, prompting state-of-the-art quantization methods to prioritize outlier mitigation while preserving model functionality. However, SLMs are typically trained through distinct methodologies that may inherently alter weight distribution characteristics. Consequently, the assumption that outlier elimination remains the primary challenge for SLM quantization lacks empirical validation, raising a critical research question: \textbf{Do SLMs exhibit fundamentally different quantization bottlenecks compared to LLMs?} Addressing this requires systematic investigation, which remains underexplored in existing literature,as shown in Figure~\ref{fig:1}.


The second critical challenge pertains to the uncertainty in translating quantization-induced efficiency gains from LLMs to SLMs. While prior studies demonstrate that quantizing LLMs can yield substantial reductions in inference latency and memory footprint~\cite{gong2024llmc}\cite{yang2024llmcbench},\cite{wang2024ptsbench} the heterogeneous architectural characteristics of SLMs may lead to divergent inference efficiency profiles post-quantization. Unlike LLMs, SLMs often employ compact designs (e.g., shallow-deep hybrid architectures or attention-linear combinations), which inherently alter their computational dependencies and memory access patterns. Consequently, quantization techniques optimized for LLMs may not proportionally benefit SLMs due to differences in their sensitivity to reduced precision arithmetic or cache utilization efficiency during inference. This raises a question: \textbf{How do quantization-induced optimizations for computational throughput and memory bandwidth interact with the unique architectural constraints of SLMs?} A systematic evaluation of quantized SLMs by different approaches is necessary to disentangle the interplay between model scale and quantization methods. Such investigations are critical to determine whether SLM quantization can achieve commensurate or superior efficiency gains relative to LLMs while maintaining task performance.


To answer aforementioned questions, in this work, we introduce SLMQuant, the first systematic benchmark for evaluating the efficacy of LLM compression techniques when applied to SLMs. Recognizing the critical research gap in SLM-specific quantization methods, we design a multi-track evaluation protocol to rigorously compare state-of-the-art quantization approaches including SmoothQuant~\cite{xiao2023smoothquant}, OmniQuant~\cite{shao2023omniquant}, and SpinQuant~\cite{liu2024spinquant} across diverse SLM architectures and downstream tasks. Our benchmark spans over 50 A800 GPU hours, encompassing 3 representative algorithms evaluated on 6 datasets and 3 model architectures. By conducting extensive ablation studies and cross-dimensional analyses, we uncover key disparities in how LLM-optimized compression methods perform on SLMs. Our empirical findings reveal critical limitations of direct LLM quantization transfer to SLMs, while identifying key factors that govern successful adaptation. These insights culminate in actionable design principles for SLM-tailored compression, bridging the gap between theoretical efficiency gains and practical deployability in edge-AI scenarios. SLMQuant thus establishes a foundational reference for advancing lightweight language models in resource-constrained applications.

Our contribution can be summarized as follows:
\begin{itemize}
    \item We construct SMLQuant, the first benchmark to comprehensively evaluate LLM compression algorithms when applying to SLMs. It provides a systematic comparison from different perspectives for the practical production of lightweight SLMs.
    \item Based on the extensive evaluation results, we conduct an in-depth analysis and provide insightful suggestions. We hope our SLMQuant can serve as the foundation to push the development of SLM compression algorithms.
\end{itemize}

\begin{figure*}
  \centering
  \includegraphics[width=0.8\linewidth]{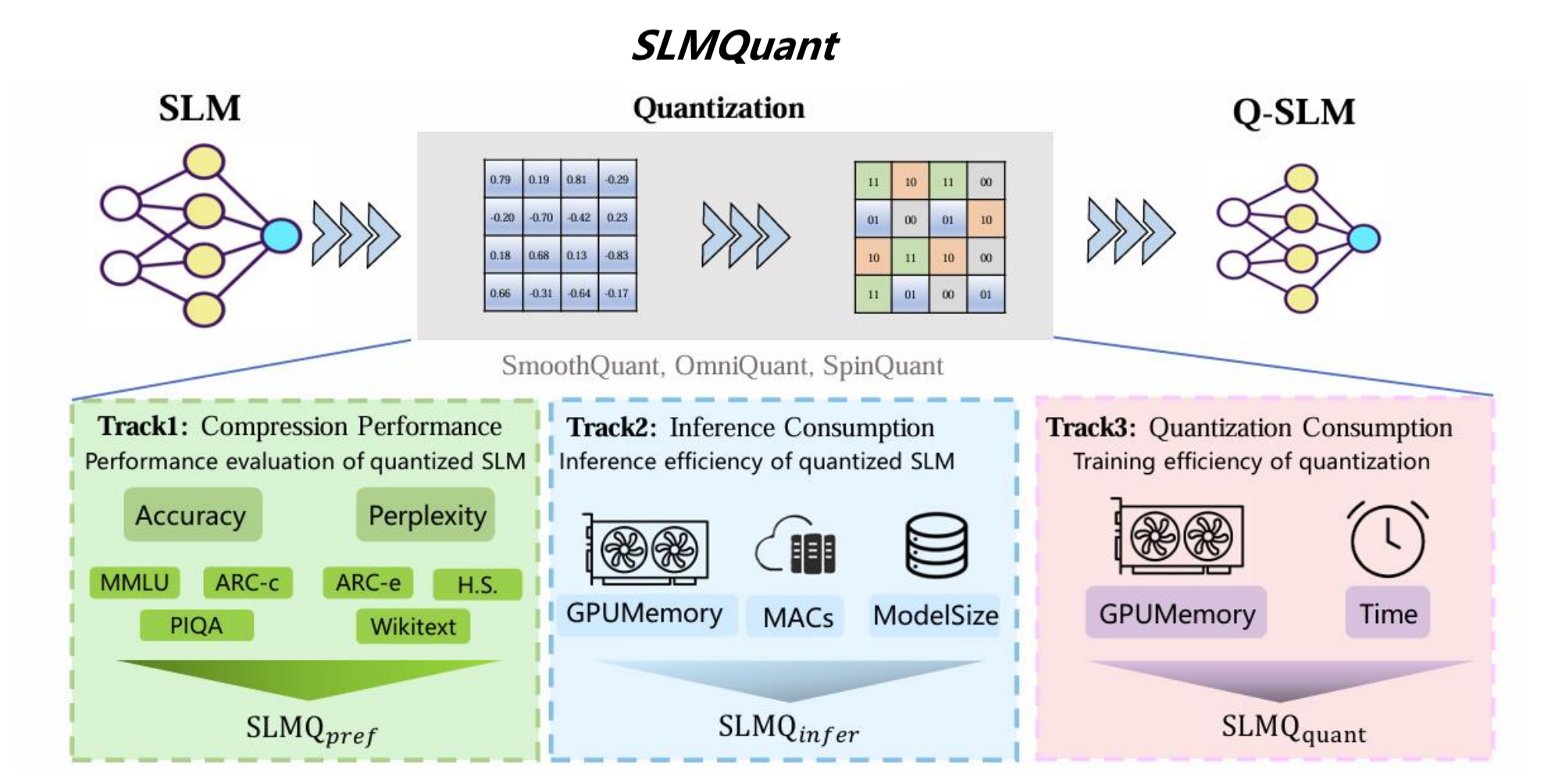} 
  \vspace{-2em}
  \caption{Overview of our SLMQuant, which consists of three evaluation tracks: compression performance, inference consumption, and quantization consumption. }
  \label{fig:2}
\end{figure*}

In summary, we introduce the first comprehensive benchmark, denoted as SLMQuant, to systematically evaluate the efficacy of diverse LLM compression algorithms in optimizing SLMs across multidimensional perspectives. We hope this work can establishes a foundational methodology for advancing future research in SLM compression, and serve as a standardized reference for guiding the development of SLM compression techniques,as shown in Figure~\ref{fig:2}.


\section{Related Works}\label{Sec:RelatedWork}
\subsection{Small Language Models}

The development of small language models (SLMs) has created new possibilities for the practical deployment of natural language processing applications. 
Compared with large language models (LLMs) whose parameter scales often reach tens of billions, SLMs typically contain millions to hundreds of millions of parameters, making them more suitable for deployment on mobile and edge devices. 
For example, the SmolLM2~\cite{allal2025smollm2} series developed by Hugging Face, with parameter sizes including 135M, 360M, and 1.7B, adopts efficient attention mechanisms to achieve outstanding performance under extremely compact model scales, thereby narrowing the performance gap between large language models and on-device applications. 
Qwen2.5-0.5B ~\cite{hui2024qwen2}, released by Alibaba Cloud as a small-scale model, balances generalization capabilities across Chinese and multilingual tasks. 
Gemini Nano ~\cite{team2023gemini}, on the other hand, is a small model launched by Google specifically for mobile devices, already deployed on Pixel devices, and demonstrates excellent inference efficiency and energy consumption performance. 
In this study, we selected several SLMs, including SmolLM2-135M, and Qwen2.5-0.5B, to construct our SLMC benchmark suite -— two models represent the cutting edge of current small model research. 
As the demand for deployment under computational constraints continues to grow, SLMs are becoming a key solution for NLP deployment, playing increasingly important roles in dialogue systems, text generation, sequence labeling, and other tasks.

\subsection{Large Language Model Quantization}

As the parameter scales of large language models (LLMs) continue to grow while the improvement in hardware performance slows down, the soaring inference costs and memory requirements have become difficult to accommodate, severely limiting their deployment and application in real-world scenarios. 
Quantization, as a classical model compression technique, can reduce model weights and activations from 32-bit floating-point to 8-bit, 4-bit, or even lower-bit integers, effectively reducing storage and computational overhead. 
For example, GPTQ ~\cite{frantar2022gptq} proposed a post-training quantization method based on layer-wise Hessian approximation, achieving efficient compression while maintaining model performance. 
SmoothQuant ~\cite{xiao2023smoothquant} significantly improves inference accuracy after activation quantization by applying scaling-based smoothing to activations. 
Building upon SmoothQuant, OmniQuant ~\cite{shao2023omniquant} optimizes smoothing factors through gradient-based methods, enabling low-bit quantization across various tasks and model architectures. 
SpinQuant ~\cite{liu2024spinquant} reduces quantization error further by applying rotation transformations and block-wise quantization to weight matrices, maintaining high model performance even at 4-bit or lower bit-widths. 
With the continuous advancement of low-bit quantization algorithms, quantization has become a key technology to enable the deployment of LLMs on servers, mobile devices, and even edge devices.

\subsection{Small Language Model Quantization}

Compared to large language models (LLMs), small language models (SLMs) have inherently smaller parameter scales, and applying quantization compression to them may lead to severe accuracy degradation. 
However, further reducing model storage and computational costs remains critical in resource-constrained scenarios. 
Unfortunately, there are currently no quantization methods specifically designed for small language models. 
The LLMC ~\cite{gong2024llmc} reports the performance of the SmolLM2 series models under various quantization methods designed for LLMs, but these methods mainly focus on optimizing large-scale models and do not fully consider the representational bottlenecks and structural characteristics of small models. 
As a result, the accuracy degradation caused by quantization on SLMs remains significant. 
On the one hand, it is necessary to comprehensively evaluate existing quantization methods on small language models; on the other hand, it is urgently required to develop quantization techniques specifically tailored for SLMs to enable their efficient deployment.

\section{Tracks and Metrics}

In this section, we present the competition tracks and evaluation metrics employed in SLMQuant, which consists of three distinct tracks. The performance metrics are designed such that higher values correspond to superior results. For clarity and interpretability, the theoretically derived scores have been scaled by a factor of 100.


\subsection{Track1: Compression Performance}

While there have been detailed comparisons of various quantization and compression methods for large language models, comprehensive evaluations for small language models remain lacking. 
In SLMQuant, we benchmark the performance of compressed small language models on mainstream evaluation datasets, focusing on their performance after quantization and compression.

To quantitatively reflect the effectiveness of different quantization algorithms, we define the following evaluation metric across all models and datasets:
\begin{equation}
\label{eq:CP}
\mathrm{SLMQ_{perf}}=\sqrt{\frac{1}{N}\sum_{i=1}^N\mathbb{E}\left(\frac{\mathbf{\textit{A}}_{i}^{c}}{\mathbf{\textit{A}}_{i}}\right)^2}.
\end{equation}
Here, $\mathbf{\textit{A}}_{i}^{c}$ and $\mathbf{\textit{A}}_{i}$ denote the accuracy scores of the compressed and pre-trained models, respectively, evaluated on the $i$th datasets. $N$ represents the total number of datasets included in the evaluation. The operator $\mathbb{E}(\cdot)$ computes the arithmetic mean across all datasets. Specifically, this formulation aggregates the average accuracy disparity between the compressed and pre-trained models over $N$ datasets, providing a holistic measure of compression efficacy across diverse evaluation scenarios. 


\subsection{Track2: Inference Performance}

The second evaluation track extends beyond conventional accuracy metrics to analyze critical inference efficiency dimensions of compressed SLMs. Specifically, we evaluate two key operational metrics: inference memory footprint and inference throughput. Memory footprint dictates deployment feasibility on resource-constrained edge devices, while inference throughput serves as a standardized metric for user-perceived responsiveness. These metrics collectively characterize the practical trade-offs between compression efficacy and real-world deployment viability.


To characterize inference efficiency in compressed SLMs, we formalize the following quantitative metric:
\begin{equation}
\mathrm{SLMQ_{infer}}=\sqrt{\frac{1}{3}\left(\mathbb{E}\left(\frac{\mathbf{\textit{M}}_{\mathrm{inf}}}{\mathbf{\textit{M}}_{\mathrm{inf}}^{c}}\right)^{2}+\mathbb{E}\left(\frac{\mathbf{\textit{TPS}}^{c}}{\mathbf{\textit{TPS}}}\right)^{2}+\mathbb{E}\left(\frac{\mathbf{\textit{S}}}{\mathbf{\textit{S}}^{c}}\right)^{2}\right)}.
\end{equation}
Here, $\mathbf{\textit{M}}_{\mathrm{inf}}$, $\mathbf{\textit{S}}$, and $\mathbf{\textit{TPS}}$ represent the peak GPU memory consumption, model storage footprint, and inference throughput (tokens-per-second) of the pre-trained SLMs during inference, respectively. $\mathbf{\textit{M}}_{\mathrm{inf}}^{c}$, $\mathbf{\textit{S}}^{c}$, and $\mathbf{\textit{TPS}}^{c}$ are those of compressed SLM at inference stage, respectively. In the mean operation $\mathbb{E}$, we calculate the mean value over all models and datasets for the corresponding metric.


\subsection{Track3: Quantization Consumption}

In addition to previous two tracks, we also formalize quantization cost as the third axis of our SLMQuant benchmark. Specifically, time efficiency is quantified through compression duration, where reduced quantization time directly correlates with lower time overhead. Concurrently, resource efficiency is operationalized via peak GPU memory consumption during compression, with diminished memory footprints indicating superior resource conservation in quantization pipelines. This dual-metric formulation enables rigorous comparative analysis of compression algorithms under practical deployment constraints.


For this track, we define the following quantitative metrics to systematically characterize algorithmic efficiency: (1) compression time, defined as the total computational time required to execute the compression pipeline, and (2) peak GPU memory consumption, measured as the maximum memory footprint observed during training stages of the quantization process:
\begin{equation}
\mathrm{SLMQ_{quant}}=\sqrt{\frac{1}{2}\left(\mathbb{E}\left(\frac{\mathbf{\textit{T}}_{\mathrm{quant}}^\mathrm{max}}{\mathbf{\textit{T}}_{\mathrm{quant}}}\right)^2+\mathbb{E}\left(\frac{\mathbf{\textit{M}}_{\mathrm{quant}}^\mathrm{max}}{\mathbf{\textit{M}}_{\mathrm{quant}}}\right)^2\right)},
\end{equation}
where $\mathbf{\textit{T}}_{\mathrm{quant}}^\mathrm{max}$ and $\mathbf{\textit{M}}_{\mathrm{quant}}^\mathrm{max}$ denote the maximal training time and peak GPU memory consumption observed across all evaluated quantization methods for a given model-dataset configuration. In this track, we calculate the mean value of training time and memory consumption over all models and datasets. To enable standardized comparisons, we normalize the metrics in this track by using the global extrema $\mathbf{\textit{T}}_{\mathrm{quant}}^\mathrm{max}$ and $\mathbf{\textit{M}}_{\mathrm{quant}}^\mathrm{max}$, which ensures higher overall metrics have better performance. 

\section{SLMQuant Implementation}

In this section, we introduce the implementation and evaluation protocal of our SLMQuant in detail.

\subsection{Implementation Details}
We implemented SLMQuant on NVIDIA A800 GPUs. Given pre-trained SLMs, we applied various compression algorithms to obtain compressed models. We use Full-Precision (FP16) to serve as the baseline. 
For SmoothQuant~\cite{xiao2023smoothquant}, we evaluate two settings: W8A8 (8-bit weights and 8-bit activations) and W4A8 (4-bit weights and 8-bit activations). Symmetric quantization for both weights and activations are used. Similarly, OmniQuant~\cite{shao2023omniquant} is evaluated under W8A8 and W4A8, but employing asymmetric weight quantization and symmetric activation quantization. SpinQuant~\cite{liu2024spinquant} is tested under the same bit settings (W8A8 and W4A8) with asymmetric quantization for both weights and activations.
In the quantization process, all the hyperparameters are the same as the open-sourced code from the original approaches.

\begin{table*}[tb]
    \caption{SmolLM-135M Compression performance of different methods.}
    \centering
    \label{tab:track1-SmolLM-135M-grouped}
    \resizebox{0.9\linewidth}{!}{
    \begin{tabular}{l|cc|ccc|ccc|c}
    \toprule
    \multirow{2}{*}{\textbf{Method}} & \multirow{2}{*}{\textbf{Model}} & \multirow{2}{*}{\textbf{Bits}} & \multicolumn{6}{c|}{\textbf{Compression Performance}} & \multirow{2}{*}{$\mathbf{SLMQ_{perf}}$} \\
    \cline{4-9}
    & & & \textbf{MMLU} & \textbf{ARC-c} & \textbf{ARC-e} & \textbf{PIQA} & \textbf{HellaSwag} & \textbf{Wiki$\downarrow$} & \\
    \midrule
    Full Prec. & SmolLM-135M & FP16 & 31.15 & 29.78 & 58.84 & 69.60 & 42.14 & 15.44 & - \\
    \midrule
    \cmidrule(lr){1-10}
    SmoothQuant & SmolLM-135M & W8A8 & 31.14 & 29.69 & 58.54 & 68.55 & 42.03 & 15.67 & 99.32\\
    OmniQuant   & SmolLM-135M & W8A8 & 31.11 & 29.10 & 58.80 & 68.06 & 41.85 & 15.89 & 98.65\\
    SpinQuant   & SmolLM-135M & W8A8 & 31.14 & 29.52 & 58.50 & 68.51 & 42.08 & 15.49 & 99.41\\
    \addlinespace[0.5em]
    \cmidrule(lr){1-10}
    SmoothQuant & SmolLM-135M & W4A8 & 27.64 & 27.30 & 46.76 & 63.33 & 36.36 & 44.64 & 81.15\\
    OmniQuant   & SmolLM-135M & W4A8 & 25.74 & 25.51 & 35.82 & 54.84 & 29.93 & 448.82 & 69.69\\
    SpinQuant   & SmolLM-135M & W4A8 & 28.28 & 27.74 & 50.76 & 67.95 & 38.42 & 17.29 & 91.45\\
    \bottomrule
    \end{tabular}%
    }
\end{table*}

\begin{table*}[tb]
    \caption{QWen2.5-0.5B Compression performance of different methods.}
    \centering
    \label{tab:track1-QWen2.5-0.5B}
    \resizebox{0.9\linewidth}{!}{
      \begin{tabular}{l|cc|ccc|ccc|c}
    \toprule
    \multirow{2}{*}{\textbf{Method}} & \multirow{2}{*}{\textbf{Model}} & \multirow{2}{*}{\textbf{Bits}} & \multicolumn{6}{c|}{\textbf{Compression Performance}} & \multirow{2}{*}{$\mathbf{SLMQ_{perf}}$} \\
    \cline{4-9}
    & & & \textbf{MMLU} & \textbf{ARC-c} & \textbf{ARC-e} & \textbf{PIQA} & \textbf{HellaSwag} & \textbf{Wiki$\downarrow$} & \\
    \midrule
    Full Prec. & QWen2.5-0.5B & FP16 & 33.46 & 32.00 & 58.33 & 70.35 & 51.29 & 13.07 & - \\
    \midrule
    \cmidrule(lr){1-10}
    SmoothQuant & QWen2.5-0.5B & W8A8 & 33.44 & 32.34 & 58.21 & 69.86 & 51.16 & 13.26 & 99.74\\
    OmniQuant   & QWen2.5-0.5B & W8A8 & 33.55 & 32.59 & 58.25 & 70.40 & 51.02 & 14.54 & 98.65\\
    SpinQuant   & QWen2.5-0.5B & W8A8 & 33.35 & 32.94 & 59.13 & 69.81 & 51.15 & 13.19 & 100.0\\
    \addlinespace[0.5em]
    \cmidrule(lr){1-10}
    SmoothQuant & QWen2.5-0.5B & W4A8 & 27.53 & 27.91 & 38.30 & 60.99 & 34.78 & 113.06 & 71.80\\
    OmniQuant   & QWen2.5-0.5B & W4A8 & 25.74 & 25.92 & 31.73 & 55.50 & 28.02 & 1614.49 & 64.09\\
    SpinQuant   & QWen2.5-0.5B & W4A8 & 31.19 & 28.75 & 50.67 & 67.52 & 44.16 & 20.54 & 86.59\\
    \bottomrule
    \end{tabular}%
    }
\end{table*}


\subsection{Evaluation Protocal}
For \textbf{track 1}, we adopt commonly used MMLU~\cite{hendrycks2020measuring}, Arc-eacy (ARC-e)~\cite{clark2018think}, and Arc-challenge (ARC-c)~\cite{clark2018think} to evaluate the knowledge ability of LLMs. 
For inference ability, we choose three datasets including PIQA~\cite{bisk2020piqa}, HellaSwag~\cite{zellers2019hellaswag}, and WikiText2~\cite{merity2016pointer} for evaluation. 
Regarding model selection, we choose two popular SLMs: SmolLM-135M~\cite{allal2025smollm2} and QWen2.5-0.5B~\cite{hui2024qwen2} for evaluation.

For \textbf{track 2}, we compare inference performance for different quantization methods. We use Wikitext2 as the measurement dataset and use both SmolLM-135M and QWen2.5-0.5B as the base model.
We set the batch size as 1 and the sequence length is set as 128.
For GPU memory measurement, we utilize the built-in tools of pytorch during inference. We use Calflops to measure the model size of SLMs before and after compression. For inference throughput, we use Calflops to report Macs and we report tokens processed per second (tokens/s), averaged over 100 iterations after 10 warm-up runs, using a fixed input sequence from Wikitext2. 

For \textbf{track 3}, we measure the resource consumption of different quantization approaches. We use Wikitext2 as the calibration dataset and report quantization costs for compressing both SmolLM-135M and QWen2.5-0.5B. For training time, We calculate the total wall-clock time required from quantization initiation to completion, including all calibration and fine-tuning procedures. For GPU memory, We measure the peak GPU memory consumption during quantization which is consistent with our inference measurement methodology to calculate $\mathrm{SLMQ_{quant}}$. 

\begin{table*}[tb]
    \caption{Inference performance of different methods.}
    \label{tab:track2-combined}
    \resizebox{0.8\linewidth}{!}{%
    \begin{tabular}{l|c|c|ccc|c}
    \toprule
    \textbf{Method} & \textbf{Model} & \textbf{Bits} & \textbf{GPU Mem.} & \textbf{Model Size} & \textbf{MACs} & $\mathbf{SLMQ_{infer}}$ \\
    \midrule
    \multirow{2}{*}{Full Prec.} 
        & SmolLM-135M & FP16 & 264.5MB & 134.52MB & 17.21G & -- \\
        & QWen2.5-0.5B & FP16 & 951.91MB & 494.03MB & 63.23G & -- \\
    \addlinespace[0.3em]
    
    \multirow{2}{*}{SmoothQuant} 
        & SmolLM-135M & W8A8 & 256.97MB & 134.52MB & 3.64G & 285.3\\
        & QWen2.5-0.5B & W8A8 & 826.24MB & 494.03MB & 17.43G & 227.2\\
    \addlinespace[0.3em]
    
    \multirow{2}{*}{OmniQuant} 
        & SmolLM-135M & W8A8 & 179.36MB & 92.29MB & 3.64G & 298.1\\
        & QWen2.5-0.5B & W8A8 & 547.49MB & 284.33MB & 17.44G & 252.9\\
    \addlinespace[0.3em]
    
    \multirow{2}{*}{SpinQuant} 
        & SmolLM-135M & W8A8 & 193.6MB & 133.24MB & 3.64G & 290.1\\
        & QWen2.5-0.5B & W8A8 & 567.67MB & 394.46MB & 17.45G & 241.5\\
    \bottomrule
    \end{tabular}%
    }
\end{table*}
\section{Experiments and Analysis}

In this section, we report the results of our experiments and provide a comprehensive analysis.

\subsection{Track 1: Compression Performance}

\textbf{8-bit quantization retains the performance of SLM.}  Table ~\ref{tab:track1-SmolLM-135M-grouped} and Table  ~\ref{tab:track1-QWen2.5-0.5B} present the results we evaluate in track 1. Quantization approaches demonstrate near-lossless preservation of accuracy, with a performance drop of less than 0.5 percent. For example, the overall metric score $\mathrm{SLMQ_{quant}}$ is over 95 percent. This indicates that 8-bit quantization is suitable for situations with no memory restrictions.

\textbf{Low-bit quantization has a significant impact on the performance of SLM.} Under 4-bit weight and 8-bit activation quantization, SpinQuant shows a significant performance deterioration on knowledge-intensive benchmarks, with MMLU scores dropping from 31.15 to 28.28 and ARC-Easy results decreasing from 58.84 to 50.76. In particular, SmoothQuant and OminQuant demonstrate even more pronounced degradation in identical quantization settings. This indicates that the quantization methods for handling outliers in large models perform poorly in the low-bit quantization process of SLM. We speculate that this might be related to the outlier distribution of SLM. 

\textbf{OmniQuant will crash under W4A8 setting.} In particular, OmniQuant demonstrate even more pronounced degradation in the W4A8 quantization settings. As a training-based method, its hyperparameters, optimized for LLMs, fail to generalize effectively to SLMs. For example, OmniQuant's parametric clipping module assumes outlier channels follow LLM-like patterns. Consequently, its LLM-optimized training framework and distribution assumptions are fundamentally misaligned with the structural characteristics of SLMs. This underscores the need for novel quantization paradigms specifically designed for SLMs. 

\textbf{Qwen has higher accuracy drop compared to SmolLM.} Despite Qwen has 3.7 times larger parameters than SmolLM under W4A8 quantization, its accuracy after compression is lower. We hypothesize that SmolLM's architecture is inherently more robust to low-bit quantization. Specifically, Qwen's standard Multi-Head Attention (MHA) and Rotary Position Embedding (RoPE) generate activations with a high dynamic range, particularly in attention scores. Low-bit quantization introduces irreversible distortion into these attention patterns, disrupting contextual relationships. In contrast, SmolLM employs NoPE (Position Embedding) and YaRN (dynamically removing rotary encoding every 4 layers), reducing floating-point dependency and minimizing positional errors post-quantization. Additionally, SmolLM utilizes Grouped-Query Attention (GQA), where grouping naturally isolates quantization noise, limiting error propagation. Experimental results confirm that architectures incorporating GQA and NoPE enable SLMs to achieve superior performance under low-bit quantization compared to those using standard MHA and RoPE.


\subsection{Track 2: Inference Performance}

\textbf{Qwen requires higher resources.}The elevated inference memory footprint of Qwen models relative to SmoLLM is primarily attributable to key architectural distinctions. Firstly, Qwen's adoption of the Layer Normalization architecture mandates the retention of larger input activations specifically for normalization computations during inference. This stands in contrast to SmolLM, which employs RMSNorm.Additionally, even when models possess comparable parameter counts, Qwen typically exhibits a higher ratio of hidden size to embedding dimension (e.g., 4:1). This structural choice directly results in larger activation tensors per layer. Moreover, it leads to significantly larger Key-Value (KV) cache volumes, an effect particularly pronounced during long-context decoding scenarios. SmolLM's more aggressively compressed dimensions (e.g., an 8:1 ratio) mitigate this specific memory demand.Collectively, these architectural differences contribute cumulatively to Qwen's greater memory demands during inference compared to SmolLM. 

\textbf{OmniQuant largely reduces GPU memory.} OmniQuant achi-eves the lowest GPU memory consumption by eliminating redundant overhead through its pure INT8 computation paradigm as shown in Table~\ref{tab:track2-combined}.Compared to SmoothQuant, its true INT8 execution flow removes the FP16 weight copies and dequantization buffers (saving 247MB for Qwen-0.5B), while against SpinQuant, its lightweight parametric scaling replaces computationally intensive rotation matrices. This reduces the memory overhead of calculating and storing rotation matrices.

\textbf{SmoothQuant still suffers from high GPU memory.} Based on the outcomes, SmoothQuant still exhibits high GPU memory requirements during inference primarily because activations must retain FP16 precision for real-time quantization operations, resulting in insignificant reduction of intermediate activation memory. Concurrently, frequent quantization/dequantization operations create barriers to operator fusion, inducing memory fragmentation. The observed memory reduction primarily reflects the benefits of weight quantization, whereas activations remain insufficiently compressed. This discrepancy between theoretical expectations and empirical results stems from inherent architectural limitations.

\begin{table*}[tb]
\centering
\caption{Quantization consumption comparison of different methods.}
\label{tab:track3-combined}
\resizebox{0.7\linewidth}{!}{%
\begin{tabular}{l|c|c|cc|c}
\toprule
\textbf{Method} & \textbf{Model} & \textbf{Bits} & \textbf{GPU Memory} & \textbf{Quant Time} & $\mathbf{SLMQ_{quant}}$ \\
\midrule
\multirow{2}{*}{Full Prec.} 
    & SmolLM-135M & FP16 & -- & -- & -- \\
    & QWen2.5-0.5B & FP16 & -- & -- & -- \\
\addlinespace[0.3em]

\multirow{2}{*}{SmoothQuant} 
    & SmolLM-135M & W8A8 & 828.31 MB & 2.02min & 549.4 \\
    & QWen2.5-0.5B & W8A8 & 2254.33MB & 2.2min & 559.6 \\
\addlinespace[0.3em]

\multirow{2}{*}{OmniQuant} 
    & SmolLM-135M & W8A8 & 2769.60 MB & 12.63min & 106.3 \\
    & QWen2.5-0.5B & W8A8 & 3713.23MB & 17.03min & 100.0 \\
\addlinespace[0.3em]

\multirow{2}{*}{SpinQuant} 
    & SmolLM-135M & W8A8 & 1558.56MB & 14.17min & 144.2 \\
    & QWen2.5-0.5B & W8A8 & 3398.21MB & 13.05min & 120.4 \\
\bottomrule
\end{tabular}%
}
\end{table*}
\begin{figure*}
  \centering
  \includegraphics[width=0.8\linewidth]{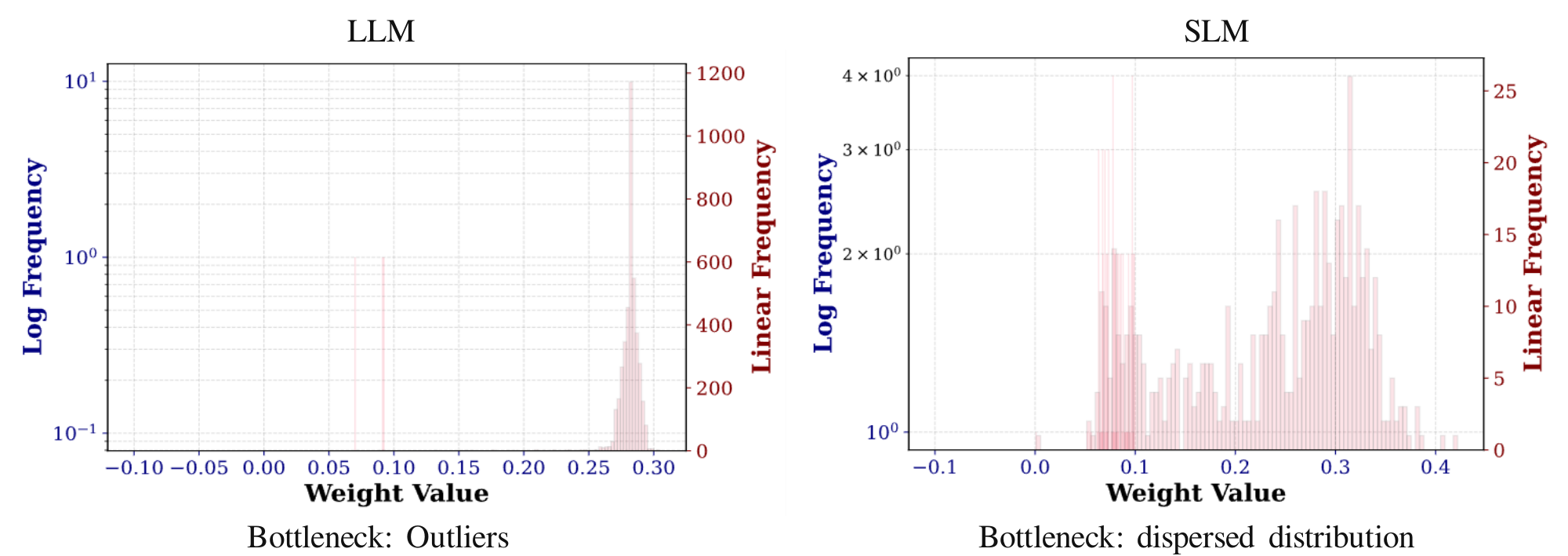} 
    \caption{The weight distributions of LLM and SLM. }
  \label{fig:4}
\end{figure*}

\subsection{Track 3: Quantization Consumption}

\textbf{Quantization time is acceptable for all the methods.} The results for track3 are shown in Table~\ref{tab:track3-combined}. The quantization durations across all methods fall within a reasonable range of 2–17 minutes. This demonstrates practical applicability in industrial deployment scenarios, where minute-scale quantization is highly efficient—particularly for offline tasks such as model pre-deployment. Compared to the hours- or days-long processes typical of model training or fine-tuning, the observed quantization times of SLMs are negligible. 

\textbf{SmoothQuant requires less quantization consumption.} \\ SmoothQuant achieves its computational efficiency through mathematically grounded equivalence transformations that strategically shift the quantization burden. This algorithmic innovation redistributes the complexity inherent in quantizing non-linear activation distributions toward the more tractable quantization of weight tensors. By resolving the core challenge of activation-value quantization through algebraic reformulation rather than empirical approximation, SmoothQuant circumvents the need for the elaborate calibration procedures that characterize alternative quantization frameworks. Specifically, comparator methodologies such as OmniQuant incur substantial overhead through their dependence on learnable parameter optimization, requiring iterative gradient-based adjustments to minimize quantization error. Similarly, SpinQuant introduces computational bottlenecks via its reliance on exhaustive iterative search mechanisms to identify optimal quantization parameters. Both approaches necessitate resource-intensive computational strategies that inherently constrain their efficiency. 

\begin{figure*}
  \centering
  \includegraphics[width=0.8\linewidth]{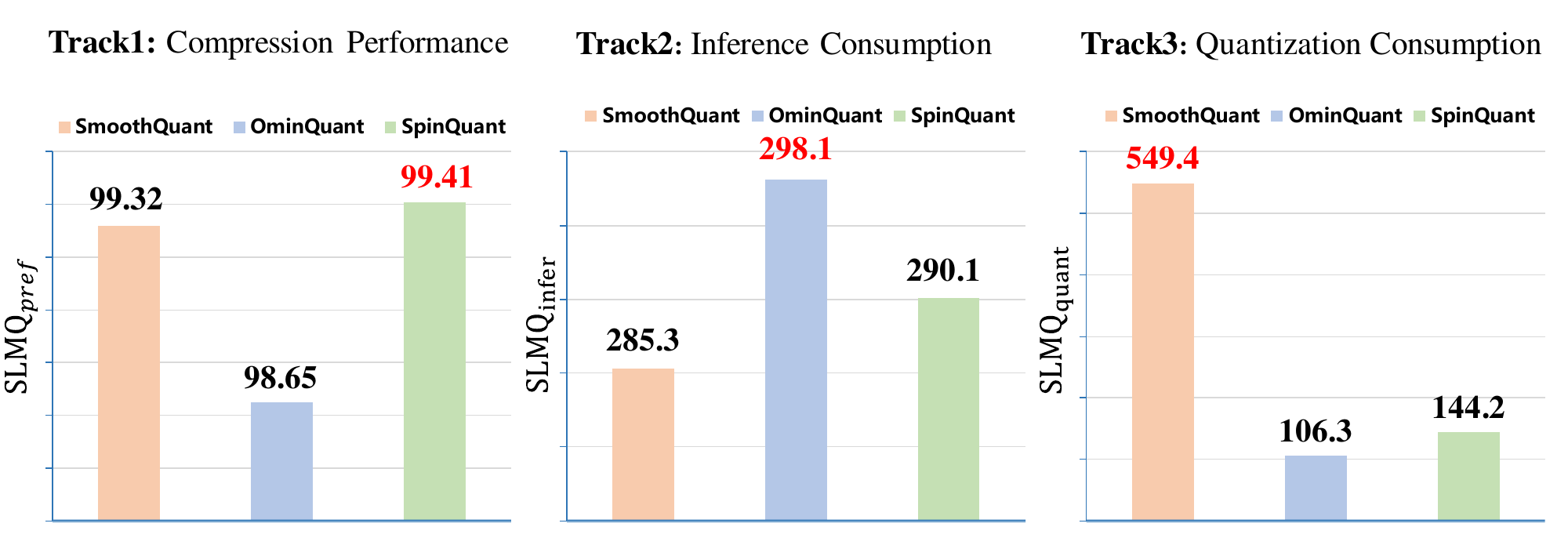} 
    \caption{Overall comparison of different quantization methods in our SLMQuant. }
  \label{fig:3}
\end{figure*}

\section{Discussion}
Based on the comprehensive evaluation of compression performance on SLMs, we draw the following conclusions : 
\begin{itemize}
    \item Our experiments demonstrate that 8-bit quantization (W8A8) consistently preserves model accuracy for SLMs, with relative performance drops generally under $0.5\%$. This aligns with prior findings on LLMs but is particularly noteworthy for SLMs, as their smaller parameter count often amplifies the effect of precision loss. The minimal degradation confirms 8-bit quantization as a safe choice for deployment scenarios where latency is primary concerns but severe memory constraints are absent.
    \item Moving to low-bit quantization, especially W4A8, causes sharp degradation in SLM performance. This is most pronounced on knowledge-intensive benchmarks. The degradation stems from the smaller representational capacity of low-bit weights, which reduces the model’s ability to maintain fine-grained semantic associations. Notably, methods like OmniQuant~\cite{shao2023omniquant}, which are optimized for LLMs, fail in the W4A8 SLM setting, due to mismatched distributional assumptions and insufficient parameter redundancy in SLMs,  as shown in figure~\ref{fig:4}. This highlights that quantization algorithms tuned for LLMs cannot be directly transplanted to SLMs without architectural and algorithmic adaptation.
    \item The robustness gap between different SLM architectures under low-bit quantization is non-trivial. Our results indicate that SmolLM exhibits better resilience than Qwen in W4A8 settings. We attribute this to three design elements: NoPE(No Positional Encoding), YaRN(Yet another RoPE extensioN) and GQA(Grouped Query Attention). By embedding absolute positional information directly into the input embeddings prior to quantization, NoPE prevents the accumulation of quantization errors inherent in calculating positional encodings within the network during inference. YaRN enhances quantization tolerance by refining RoPE's wavelength scaling mechanism, resulting in a smoother numerical distribution of positional encodings that is less sensitive to low-precision perturbation. GQA's parameter sharing across attention heads constrains head diversity, centralizing parameter distributions. This structural regularization demonstrably reduces activation tensor dynamic range by 2-3x, significantly mitigating quantization perturbation. This suggests that quantization-aware architectural design, traditionally overlooked in SLMs, may represent a promising research direction, with significant potential for co-design approaches that incorporate quantization constraints into the architecture search process.
    \item From the results in figure~\ref{fig:3}, we observe that training-free quantization methods like SmoothQuant typically exhibit minimal computational overhead during the quantization process. Conversely, OmniQuant significantly reduces inference overhead as it requires no additional parameters during inference. SpinQuant, however, adopts a more resource-intensive approach, trading increased computational cost for enhanced model accuracy. 
    \item Prompt engineering exhibits markedly greater criticality for SLMs compared to LLMs. The inherent limitations of SLMs render them highly susceptible to prompt quality; suboptimal or poorly structured prompts readily induce task misalignment or catastrophic failure. In stark contrast, the robustness derived from the extensive pre-training and scale of LLMs significantly dampens their sensitivity to prompt variations. A compelling illustration of this phenomenon is observed on knowledge-intensive benchmarks like MMLU. Consider the zero-shot performance of Llama2-7B  versus smolLM2-135MB. Without explicit task instruction or context in the prompt, Llama2-7B maintains a functional baseline performance on MMLU. While significantly lower than its potential with optimized prompting, its performance does not collapse entirely; it retains some ability to parse questions and access relevant knowledge.Conversely, smolLM2-135MB exhibits near-complete breakdown under the same zero-shot condition.This stark contrast underscores the absolute necessity of precise prompt engineering for viable SLM application on challenging tasks.
\end{itemize}

\section{Conclusion}
In this paper, we presented a Small Language Model Quantization Benchmark (SLMQuant) to systemically evaluate the SLM quantization algorithms. Based on the evaluation results, we also provide an in-depth analysis to guide the further design of SLM quantization approaches. We hope our SLMQuant can contribute insightful suggestions and serve as a foundation for future research. 
One limitation of our SLMQuant is that we only choose the three most representative approaches. We will include more SLM quantization algorithms, such as SLM KV cache quantization, in our future work. We will also introduce more tracks and datasets, such as coding datasets and mathematical datasets, to conduct more comprehensive tests on the quantized SLMs.

\begin{acks}
This work was supported by the Beijing Municipal Science and Technology Project (No. Z231100010323002), the National Natural Science Foundation of China (Nos. 62306025, 92367204) , CCF-Baidu Open Fund.
\end{acks}

\bibliographystyle{ACM-Reference-Format}
\bibliography{sample-base}


\end{document}